\documentclass[reqno, 10pt]{amsart}

\usepackage{amsmath,amssymb,amsthm, booktabs}

\usepackage{tikz}
\usepackage{caption}
\usepackage{graphicx}
\usepackage{graphics}
\usepackage{tikz}	
\usepackage{algorithm}
\usepackage{algorithmicx}
\usepackage{algpseudocode}
\usepackage{url}

\begin{document}

\title[]{An effective variant of the \\Hartigan k-means algorithm}

\author[]{Fran\c{c}ois Cl\'ement}
\address{Department of Mathematics, University of Washington, Seattle}
\email{fclement@uw.edu}

\author[]{Stefan Steinerberger}
\address{Department of Mathematics and Department of Applied Mathematics, University of Washington, Seattle}
\email{steinerb@uw.edu}

\begin{abstract}
The k-means problem is perhaps \textit{the} classical clustering problem and often synonymous with Lloyd's algorithm (1957). It has become clear that Hartigan's algorithm (1975) gives better results in almost all cases, Telgarsky-Vattani note a typical improvement of $5\%$ -- $10\%$. We point out that a very minor variation of Hartigan's method leads to another $2\%$ -- $5\%$ improvement; the improvement tends to become larger when either dimension or $k$ increase.
\end{abstract}

\maketitle

\section{Introduction}
\subsection{k-means} The $k-$means problem is as follows: given $x_1, \dots, x_n  \in \mathbb{R}^d$ and $k \in \mathbb{N}$, the goal is to partition the $n$ points into $k$ clusters $S_1, \dots, S_k$ such that
$$ \sum_{i=1}^{k} \sum_{x \in S_i} \left\| x - \frac{1}{|S_i|} \sum_{j \in S_i} x_j\right\|^2 \rightarrow \min.$$
The center of mass $|S_i|^{-1} \sum_{j \in S_i} x_j$ is also sometimes called the centroid $\mu_i \in \mathbb{R}^d$. Geometrically, we are asked to partition the set into $k$ sets that are all as close as possible to their center of mass. This problem has a long history: it was proposed, independently, by Steinhaus \cite{steinhaus} in 1956, Lloyd \cite{lloyd} in 1957, Ball and Hall \cite{ball} in 1965 and MacQueen \cite{macqueen} in 1967 (see Jain \cite{jain}). The problem is NP-hard even for $k=2$ clusters \cite{aloise}, the best one can hope for are approximations to the true solution.

\begin{center}
\begin{figure}[h!]
        \begin{tikzpicture}
        \node at (1,0) {\includegraphics[width=0.22\textwidth]{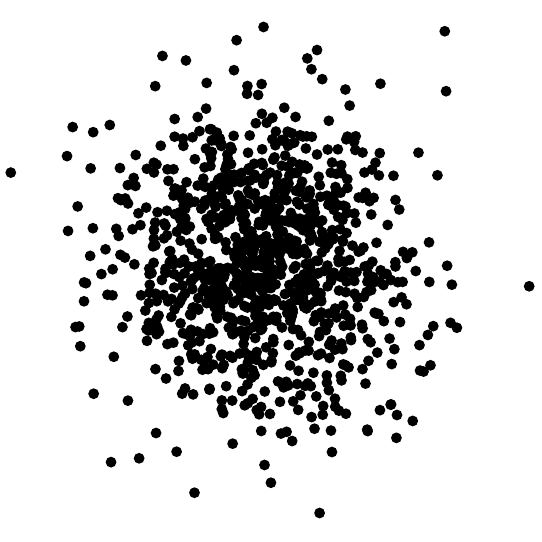}};
        \node at (3.5,0) {\LARGE $\implies$};
     \node at (6,0) {\includegraphics[width=0.22\textwidth]{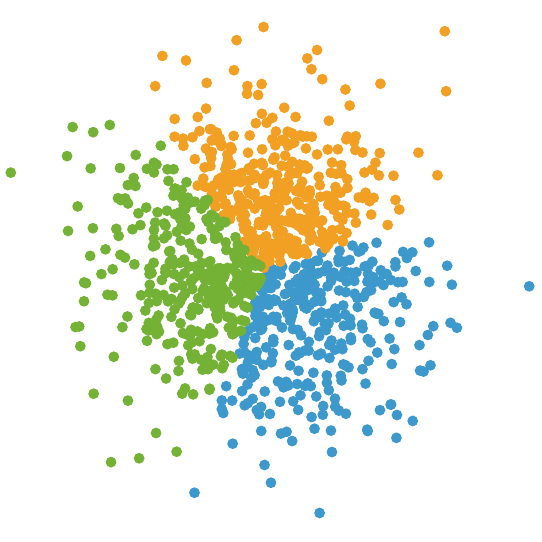}};
    \end{tikzpicture}
    \caption{Thousands points in $\mathbb{R}^2$ are clustered into $k=3$ sets.}
\end{figure}
\end{center}

 \subsection{Lloyd's algorithm.}
The most commonly used way to find an approximate solution is the algorithm proposed by Lloyd (and, independently, Forgy \cite{forgy}), with a variant by MacQueen~\cite{macqueen}. For any given cluster partition $S_1, \dots, S_k$ with associated centroids $\mu_1, \dots, \mu_k$, the algorithm goes through every point and assigns $x_j$ to the cluster $S_{\ell}$ whose centroid $\mu_{\ell}$ is closest to $x_j$ (breaking ties in an arbitrary manner). After that, the centroids are updated to account for new cluster assignments and the procedure is repeated. The $k-$means functional can only decrease under one step of this algorithm, the algorithm stops when no point is moved anymore. Since the functional is monotonically decreasing and bounded from below, the algorithm stops eventually. Lloyd's algorithm is almost trivial to implement, easy to explain and still very widely used today (for example in \textsc{Python}'s sklearn.cluster.KMeans).

 \subsection{Hartigan's algorithm.}
Hartigan's algorithm \cite{hartigan} was proposed in 1975. It seems to have not been considered much until the early 2010's when it was again popularized by Telgarsky-Vattani \cite{telgarsky}, Slonim-Aharoni-Crammer \cite{slonim} and others. The algorithm has an easy motivation: suppose $x$ is a point currently assigned to cluster 1 and $\|x - \mu_1\| = \|x - \mu_2\|$. Then Lloyd's algorithm would be indifferent about moving $x$, it is already connected to a centroid of closest distance. However, note that if we were to move $x$ over the cluster 2, then $\mu_2$ would move in the direction of $x$ since $x$ would then be factored into how $\mu_2$ is computed
and the k-means functional would decrease.

\begin{center}
\begin{figure}[h!]
        \begin{tikzpicture}
    \filldraw (0,0) circle (0.07cm);
    \node at (0, -0.3) {$\mu_1$};
     \filldraw (4,0) circle (0.07cm);
   \node at (4, -0.3) {$\mu_2$};
   \filldraw (2,0) circle (0.07cm);
      \node at (2, -0.3) {$x$};
   \draw [dashed] (0,0) -- (2,0);
    \end{tikzpicture}
    \caption{A point $x$ assigned to cluster 1 but having the same distance to $\mu_2$ that it has from $\mu_1$.}
\end{figure}
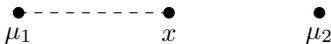
\end{center}
\vspace{-10pt}
A more formal explanation is as follows (see also \S 2 for pseudocode): if $S_i$ is a cluster, then we abbreviate its contribution to the $k-$means functional via
$$ \phi(S_i) = \sum_{x \in S_i}\left\| x- \frac{1}{|S_i|} \sum_{y \in S_i}^{|S_i|} y \right\|^2=   \sum_{x \in S_i}\left\| x- \mu(S_i) \right\|^2.$$
Suppose now that $x \in S_i$. Then the \textit{decrease} in the $k-$means functional when moving $x$ from $S_i$ to $S_j$ is given by
$$ \Delta(x, S_i, S_j) = \phi(S_i) + \phi(S_j) - \phi(S_i \setminus \left\{x\right\}) - \phi(S_j \cup \left\{x \right\})$$
and a bit of algebra shows that this can be rewritten as
$$\Delta(x, S_i, S_j) = \frac{|S_j|}{|S_j| + 1} \| \mu(S_j) - x\|^2 - \frac{|S_i|}{|S_i|-1}\| \mu(S_i) - x\|^2.$$
Hartigan's method now picks a point $x \in S_i$, checks whether there exists $j \neq i$ such that $\Delta(x, S_i, S_j) > 0$ and, if so, then it removes $x$ from $S_i$ and assigns it to the cluster indexed by $\arg\max_{\ell} \Delta(x, S_i, S_{\ell})$.
As pointed out by Telgarsky-Vattani \cite{telgarsky}, every local minimum of Hartigan's algorithm is also a local minimum of Lloyd's algorithm but not necessarily the other way around: Hartigan's algorithm may further improve a local Lloyd-minimum. Telgarsky-Vattani \cite{telgarsky} summarize an empirical comparison by saying that on \textit{average, Hartigan’s method provides an improvement of roughly 5-10\%} over Lloyd's method. It is becoming more popular: $\textsc{kmeans}()$ in \textsc{R} calls the Hartigan-Wong implementation \cite{hw}.

\section{A variation of the Hartigan algorithm}
\subsection{The Algorithm.} We now present a simple variation of Hartigan's method.\footnote{The authors kept referring to it, tongue-in-cheek, as Smartigan (because it is both a good idea and also extremely close to Hartigan's method), slowly got used to the nickname and now cannot bear to part with it.}
We describe a particular formulation of Hartigan's method (where points are evaluated in the order given by a random permutation which slightly outperforms iid random sampling); our new algorithm is identical except in a single line where the difference is made explicit.

\begin{algorithm}[h!]
\caption{Hartigan/Smartigan Algorithm. }
\begin{algorithmic}
    \Require A set $\left\{x_1, \dots, x_n \right\} \subset \mathbb{R}^d$, number of clusters $k$, max iterations $N_{\mbox{\tiny iter}} \in \mathbb{N}$
    \State Initialize cluster $C_1,\ldots,C_k$ in some way
    \State Change=False
    \State $n_{\mbox{\tiny iter}}=0$
    \While{Change=False and $n_{\mbox{\tiny iter}}<N_{\tiny \mbox{iter}}$} 
        \State Take a random permutation $\pi$ of $\{1,\ldots,n\}$
        \State Change $=$ True
        \For{$i=1$ to $n$} 
        \State Consider the point $x_{\pi(i)}$, currently associated with cluster $C_r$.
        \If {$|C_r|=1$}
        \State Pass
        \Else
            \State Find cluster $C_j$ (different from $C_r$)  minimizing 
            $$\Delta_j=\frac{|C_j|}{|C_j|+1}||x_{\pi(i)}-\mu(C_j)||^2$$
            \If {$$\Delta_j\leq \frac{|C_r|}{|C_r|-1}||x_{\pi(i)}-\mu(C_r)||^2 \cdot \begin{cases} 
1 \qquad &\mbox{(Hartigan)}\\
\frac{3}{2} - \frac{1}{2}\frac{n_{\mbox{\tiny iter}}}{N_{\mbox{\tiny iter}}} \qquad &\mbox{(Smartigan)}
            \end{cases}
            $$}
                \State Assign $x_{\pi(i)}$ to $C_j$
                \State Update $\mu(C_j)$ and $\mu(C_r)$
                \State Change $=$ False
            \EndIf
        \EndIf
        \EndFor
        \State $n_{{\mbox{\tiny iter}}}=n_{\mbox{\tiny iter}}+1$
    \EndWhile
    \State Return sum of squared distances
\end{algorithmic}
\end{algorithm}

\subsection{Remarks}
 Several remarks are in order.
\begin{enumerate}
\item Hartigan's method is the obvious thing once one already is close to a good clustering.  However, in the beginning, one may only have a very vague notion of the underlying cluster structure: Smartigan encourages exploration.
\item Smartigan, asymptotically, turns into Hartigan; one might specify the algorithm to run Hartigan at the very end which would ensure that one inherits all the guarantees that one has for a Hartigan minimizer.
\item The choice of constants in $3/2 - n_{\mbox{\tiny iter}}/(2N_{\mbox{\tiny iter}})$ is motivated by experiments; the main idea suggests that one should choose a monotonically decreasing function in $n_{\mbox{\tiny iter}}$ that approaches 1 as $n_{\mbox{\tiny iter}}$ approaches $N_{\mbox{\tiny iter}}$. Many such functions are conceivable and many seem to lead to good improvements; we picked the linear function for the sake of concreteness, simplicity and performance in practice.
\item When it comes to actual performance, there are relatively few theoretical results in the literature; the difference between Lloyd's algorithm and Hartigan's method is seen through numerical experimentation. Moreover, see Telgarsky-Vattani \cite{telgarsky}, the supremacy of Hartigan's method is not subtle but very clear and easily observable. Likewise, we will argue that Smartigan outperforms Hartigan in a manner that is equally clear (but smaller in scale than the Lloyd $\rightarrow$ Hartigan improvement). 
\end{enumerate}

\subsection{Theoretical guarantees.} 
Very few things are rigorously known for any of these algorithms.  It is easy to see that Lloyd-stable assignment, a cluster assignment that remains unchanged under Lloyd's algorithm, is the weakest form of guarantee: every Hartigan-stable configuration is also Lloyd-stable (since Hartigan is more prone to changing cluster assignments). Similarly, we may deduce that any Smartigan-stable configuration is Hartigan-stable and thus Lloyd-stable. 

\begin{center}
    \begin{tikzpicture}
        \node at (0,0) {Smartigan-stable};
        \node at (2,0) {$\subseteq$};
                \node at (4,0) {Hartigan-stable};
        \node at (6,0) {$\subseteq$};
        \node at (8,0) {Lloyd-stable};
    \end{tikzpicture}
\end{center}

 We emphasize that Smartigan-stability, the guarantee that cluster assignments remain unchanged independently of how many times the Smartigan algorithm is applied, can be seen as a very powerful form of Hartigan-stability (with an additional safety margin of $50\%$). One may artificially define Smartigan$^*$ as Smartigan followed by Hartigan in which case one trivially recovers the guarantees of Hartigan's algorithm -- this, however, would be missing the main point which is the greater exploration of configuration space that occurs early on.

\subsection{Un commentaire sociologique.} There is a curious discrepancy in the literature that deserves a short sociological comment. The importance of $k-$means in the literature is beyond doubt; however, there is a gap between how well-known the $k-$means problem is and by how much Lloyd's algorithm is reliably and substantially outperformed by Hartigan's method. This is sometimes, but rarely, hinted at in the literature. Slonim-Aharoni-Crammer mention that \textit{the complexity of both algorithms is similar, and since both are equally trivial to implement, one might wonder why is it that Lloyd’s algorithm is so prevalent while Hartigan’s algorithm is scarcely used in practice} \cite{slonim}. We have no explanation. It is conceivable that the simplicity of Lloyd's algorithm, it being taught at a basic level, and its easily available implementations give it a distinguished position in the literature that is never questioned. We want to emphasize that the improved performance of both Hartigan and then the further improvement by Smartigan, both easily validated, suggest that it is conceivable that \textit{the problem of} \underline{actually} \textit{minimizing the $k-$means functional may have never received the attention it deserves}.

\section{Numerical Results}

\subsection{Real Data: Low Dimensions}
We start with a classic example: the \textit{Fisher Iris Data Set} comprised of 150 points in $\mathbb{R}^4$ describing three different subtypes of the flower Iris, each of them represented 50 times. It is known to be an imperfect example which $k-$means will not solve with perfect accuracy when $k=3$.

\begin{table}[h]
\centering
\label{tab:iris_kmeans}
\begin{tabular}{@{}lccccc@{}}
\textbf{Algorithm} & $k=3$ & $k=4$ & $k=5$ & $k=10$ & $k=20$   \\ \midrule
Lloyd          &   96.88           &   80.82       & 77.02    &   55.61   &  21.68   \\
Hartigan       &  \textbf{78.85}   &   \textbf{57.97}      & 51.37     &   29.86  &  17.38  \\
Smartigan      &     79.23         &   59.16          &  \textbf{49.28}  &  \textbf{28.18}  & \textbf{16.76}         \\ \bottomrule
\end{tabular}
\caption{Average performance on the Fisher Iris Dataset (random initialization, averaged over 500 runs each)}
\end{table}

 Another reasonably generic data set is \textit{Fisher's cat data set}; for each of the 144 cats, the gender, total weight (in kg) and weight of the heart (in g) is recorded. We dropped the gender and worked with the remaining data. The picture is again quite consistent: for $k=2$, there seems to be no difference between Hartigan and Smartigan, there is a mild improvement for larger values of $k$ and substantial improvements for $k=10$ and $k=20$.

\begin{table}[h]
\centering
\label{tab:iris_kmeans}
\begin{tabular}{@{}lccccc@{}}
\textbf{Algorithm} & $k=2$ & $k=3$ & $k=5$ & $k=10$ & $k=20$   \\ \midrule
Lloyd          &   351.05           &   228.472       & 163.07    &   64.37   &  34.49   \\
Hartigan       &  \textbf{309.128}   &   184.231      & 83.89    &   34.10  &  19.97  \\
Smartigan      &  \textbf{309.128}  &   \textbf{182.738}  &  \textbf{83.86}  &  \textbf{31.00}  & \textbf{16.94}         \\ \bottomrule
\end{tabular}
\caption{Average performance on Fisher's Cat Dataset (random initialization, averaged over 500 runs each)}
\end{table}

\subsection{Real Data: High Dimensions}
 The next example is the \textit{Breast Cancer Wisconsin} dataset \cite{wis} containing 569 points in $\mathbb{R}^{30}$ (30 features of 569 tumors).

\begin{table}[h]
\centering
\label{tab:iris_kmeans}
\begin{tabular}{@{}lccccc@{}}
\textbf{Algorithm} & $k=2$ & $k=3$ & $k=5$ & $k=10$ & $k=20$   \\ \midrule
Lloyd          &   12.93           &   8.08      & 6.56    &   5.42   &  3.46   \\
Hartigan       &  \textbf{7.79}   &   5.05      & 2.12   &   1.12  &  0.88  \\
Smartigan      &  \textbf{7.79}  &   \textbf{4.95}  &  \textbf{2.06}  &  \textbf{1.06}  & \textbf{0.86}         \\ \bottomrule
\end{tabular}
\caption{Average performance on BCW  (random initialization, averaged over 100 runs each; all numbers $\times 10^{7}$).}
\end{table}

The final examples come from Lederman et al.~\cite{lederman}. We took the \emph{exact} implementation and test cases \cite{royexact}, and modified \emph{precisely} three lines of code: the cluster change condition on which Smartigan differs from Hartigan, and the function definition to include $n_{\mbox{\tiny iter}}$. This allows us to replicate perfectly their results from 4 datasets presented in Table 1 in~\cite{lederman}: the Olivetti faces dataset~\cite{samaria}, and three datasets from the 20 newsgroups dataset~\cite{mitchell}. Table~\ref{tab:copy} shows that Smartigan gives comparable results for the $k$-means loss, with usually better NMI values (correlation between the output clustering and the true labels, the closer to 1 the better). Lederman et al.~\cite{lederman} also compared the Hartigan method to the SDP algorithm of Peng-Wei \cite{peng} as well as the spectral clustering method of Shi-Malik \cite{shi} with Hartigan leading to superior results both in terms of the functional and NMI, we omit these results for the sake of brevity.

\begin{table*}[h]
    \vskip -0.1in
    \begin{center}
        \begin{small}
            \begin{sc}  
                \begin{tabular}{lccc|cc|cc}
                    \multicolumn{4}{c}{Dataset Parameters} & \multicolumn{2}{c}{$k$-Means loss} & \multicolumn{2}{c}{NMI} \\
                     & $n$ & $d$ & $K$  & Hartigan & Smartigan   & Hartigan & Smartigan  \\
                    \midrule
                    Olivetti & 400 & 4096 & 40  & 8.11 & \bf{8.00}  & 0.77 & \bf{0.78} \\
                    20NG-A & 200  & 5000 & 2  & \bf{193.46} & \bf{193.46}  & 0.54 & \bf{0.62}\\
                    20NG-B & 500  & 5000 & 5  & \bf{481.72} & 481.90 & 0.44 & \bf{0.49} \\
                    20NG-C & 1000  & 5000 & 10 & \bf{951.96} & 953.41 & \bf{0.31} & 0.25 \\
                    \bottomrule
                \end{tabular}
                    \caption{Results obtained by taking the implementation from~\cite{lederman} and changing 3 lines to obtain Smartigan. }
                      \label{tab:copy}
            \end{sc}
        \end{small}
    \end{center}
\end{table*}

\subsection{Synthetic Data Sets.}
It is easy to generate synthetic data. We consider two different types of examples.
\begin{enumerate}
    \item In the small distance examples, we sample $k_s$ centers from $[0,3]^d$ and define them to be Gaussians with covariance matrix $0.3\cdot \mbox{Id}_{d \times d}$.
    \item In the large distance examples, we sample $k_{l}$ centers from $[0,5]^d$
    and consider Gaussians with covariance matrix $0.1\cdot \mbox{Id}_{d \times d}$.
\end{enumerate}
 The small distance problem is more complicated than the large distance problem.
  Clusters are not guaranteed to have the same number of points, we assign a point to a random cluster before generating the point. 

\begin{table}[h]
\centering
\label{tab:2d_kmeans_many}
\begin{tabular}{@{}lcccccc@{}}
$d=2$ & $k_s=2$ &  $k_s=10$ & $k_s=25$& $k_{l}=2$  &$k_l=10$& $k_l=25$  \\ \midrule
$n=250$        &   $>-0.1\%$          &  {\boldmath  $-2.3\%$ }     & {\boldmath  $-4.4\%$} & $>-0.1\%$ & {\boldmath  $-4.2\%$} & {\boldmath  $-4.3\%$}\\
$n=500$     &  $>-0.1\%$   &  {\boldmath   $-1.5\%$}      & {\boldmath  $-2.9\%$} &$>-0.1\%$ & {\boldmath  $-2.7\%$} & {\boldmath  $-3.0\%$}\\
$n=1\,000$      &  $>-0.1\%$  &  {\boldmath   $-1.2\%$}  &  {\boldmath  $-2.1\%$} & $>-0.1\%$ & {\boldmath   $-2.7\%$} & {\boldmath  $-2.1\%$}\\\bottomrule
\end{tabular}
\caption{Average difference between Hartigan/Smartigan in 2 dimensions  (100 different point sets, k-means++ initialization and averaged over 20 runs each). Negative means Smartigan is better, while $>0.1\%$ indicates the difference is negligible.}
\end{table}

Any random element of the algorithm is done in the same way for both Hartigan and Smartigan in all these tests: they start with the same point sets and cluster assignments, and the order in which the points are considered is exactly the same. 
The difference between the two algorithms is already quite noteworthy in two dimensions. When there are only 2 clusters, the problem is easy enough to be solved by either method and we do not see a measurable difference between the two methods (they basically both solve the problem perfectly). However, as soon as there are more clusters, Smartigan leads to consistently better results.

\begin{table}[h!]
\centering
\label{tab:5d_kmeans_many}
\begin{tabular}{@{}lcccccc@{}}
$d=5$ & $k_s=2$ &  $k_s=10$ & $k_s=25$& $k_{l}=2$  &$k_l=10$& $k_l=25$  \\ \midrule
$n=250$        &   $<0.1\%$          &  {\boldmath $-1.7\%$}      & {\boldmath$-2.2\%$} & $<0.1\%$ &{\boldmath$-11.8\%$} & {\boldmath $-9.3\%$}\\
$n=500$     &  $<0.1\%$   &  {\boldmath $-1.3\%$  }    &  {\boldmath $-1.6\%$} & $<0.1\%$ & {\boldmath$-8.6\%$} & {\boldmath$-8.4\%$}\\
$n=1\,000$      &  $<0.1\%$  &   {\boldmath $-0.9\%$}  &  {\boldmath $-1.2\%$} & $<0.1\%$ & {\boldmath $-8.4\%$} & {\boldmath $-7.1\%$} \\\bottomrule
\end{tabular}
\caption{Average percentage difference between Hartigan and Smartigan in 5 dimensions  (100 different point sets, each with k-means++ initialization and averaged over 20 runs each).}
\end{table}

The results remain consistent in higher dimensions with the actual improvements becoming larger and larger. In all three cases, dimension $d=2$, dimension $d=5$ and dimension $d=20$, the case of two clusters is solved in a way leading to very comparable scores by both methods; the moment the number of clusters increases, Smartigan gains a definitive advantage.

\begin{table}[h]
\centering
\label{tab:20d_kmeans_many}
\begin{tabular}{@{}lcccccc@{}}
$d=20$ & $k_s=2$ &  $k_s=10$ & $k_s=25$& $k_{l}=2$  &$k_l=10$& $k_l=25$  \\ \midrule
$n=250$        &   $0\%$          &  {\boldmath $-6.3\%$}      & {\boldmath $-5.5\%$} & $0\%$ & {\boldmath$-16.0\%$} & {\boldmath $-24.1\%$}\\
$n=500$     &  $0\%$   &   {\boldmath $-6.1\%$}      &{\boldmath $-4.8\%$} & $0\%$ & {\boldmath$-12.3\%$} & {\boldmath $-19.6\%$}\\
$n=1\,000$      &  $0\%$  &  {\boldmath $-5.2\%$}  & {\boldmath $-4.3\%$} & $0\%$ & {\boldmath$-9.8\%$} & {\boldmath$-17.4\%$}\\\bottomrule
\end{tabular}
\caption{Average percentage difference between Hartigan and Smartigan in 20 dimensions  (100 different point sets, each with k-means++ initialization and averaged over 20 runs each).}
\end{table}

\textbf{Acknowledgment.} We are grateful to Roy Lederman for insightful discussions.

\end{document}